\documentclass[journal]{IEEEtran}
\usepackage{xspace} 
\usepackage{amsmath} 
\usepackage{graphicx}
\usepackage{booktabs}   
\usepackage{multirow}   
\usepackage{makecell}   
\usepackage{caption}
\usepackage{xcolor}
\usepackage{float}
\usepackage[margin=1in]{geometry}

\usepackage{makecell}
\usepackage{multirow}
\usepackage[table]{xcolor}
\usepackage[normalem]{ulem}
\newcommand{\best}[1]{\textbf{#1}}
\newcommand{\second}[1]{\uline{#1}} 
\definecolor{oursblue}{RGB}{221,235,247} 
\newcommand{\na}{\textcolor{black!45}{--}}
\newcommand{\BlueHL}[1]{\colorbox{oursblue}{\hspace{0.15em}#1\hspace{0.15em}}}
\usepackage{bbm}
\usepackage{graphicx}  
\usepackage{subcaption}
\usepackage{amssymb}

\makeatletter
\DeclareRobustCommand\onedot{\futurelet\@let@token\@onedot}
\def\@onedot{\ifx\@let@token.\else.\null\fi\xspace}

\makeatother

\usepackage{amsmath,amsfonts}
\usepackage{algorithm}
\usepackage{algorithmic}
\usepackage{array}
\usepackage[caption=false,font=normalsize,labelfont=sf,textfont=sf]{subfig}
\usepackage{textcomp}
\usepackage{stfloats}
\usepackage{url}
\usepackage{verbatim}
\begin{document}

\title{Evaluation as Evolution: Transforming Adversarial Diffusion into Closed-Loop Curricula for Autonomous Vehicles}

\author{Yicheng Guo,
        Jiaqi Liu,
        Chengkai Xu,
        Peng Hang,~\IEEEmembership{Senior Member,~IEEE,}
        Jian Sun

\thanks{Yicheng Guo, Chengkai Xu, Peng Hang, and Jian Sun are with the College of Transportation, Tongji University, Shanghai 201804, China. (e-mail: 2410796@tongji.edu.cn, xck1270157991@gmail.com, hangpeng@tongji.edu.cn, sunjian@tongji.edu.cn)}

\thanks{Jiaqi Liu is with the Department of Computer Science, University of North Carolina at Chapel Hill, United States. (e-mail: jqliu@cs.unc.edu)}

\thanks{Corresponding author: Peng Hang}
}

\maketitle

\begin{abstract}
Autonomous vehicles in interactive traffic environments are often limited by the scarcity of safety-critical tail events in static datasets, which biases learned policies toward average-case behaviors and reduces robustness.
Existing evaluation methods attempt to address this through adversarial stress testing, but are predominantly open-loop and post-hoc, making it difficult to incorporate discovered failures back into the training process.
We introduce Evaluation as Evolution ($E^2$), a closed-loop framework that transforms adversarial generation from a static validation step into an adaptive evolutionary curriculum. Specifically, $E^2$ formulates adversarial scenario synthesis as transport-regularized sparse control over a learned reverse-time SDE prior. To make this high-dimensional generation tractable, we utilize topology-driven support selection to identify critical interacting agents, and introduce Topological Anchoring to stabilize the process. This approach enables the targeted discovery of failure cases while strictly constraining deviations from realistic data distributions. 
Empirically, $E^2$ improves collision failure discovery by 9.01\% on the nuScenes dataset and up to 21.43\% on the nuPlan dataset over the strongest baselines, while maintaining low invalidity and high realism. It further yields substantial robustness gains when the resulting boundary cases are recycled for closed-loop policy fine-tuning.
\end{abstract}


\section{Introduction}

\IEEEPARstart{A}{utonomous} driving systems increasingly operate in open, interactive environments where decisions emerge from continuous feedback between the agent and its surroundings~\cite{liu2025aligning, han2025multimodal,liu2025agent0}.
While supervised and imitation learning on large-scale logs have advanced rapidly, the resulting policies often remain biased toward average-case behaviors~\cite{wu2024recent}.
However, it is the tail that defines safety risk in practice: rare, high-stakes, multi-agent events where small perturbations are amplified through closed-loop feedback~\cite{liu2026risknet}.

Static datasets are sparsest precisely where the decision boundary is most fragile, offering limited counterfactual coverage for recovering under strategic, co-adapting traffic. Consequently, the bottleneck is not merely data quantity, but the lack of boundary experience that concentrates learning signals on failure-prone regions of the dynamics.

\begin{figure}[t]
  \centering
  \includegraphics[width=0.95\linewidth]{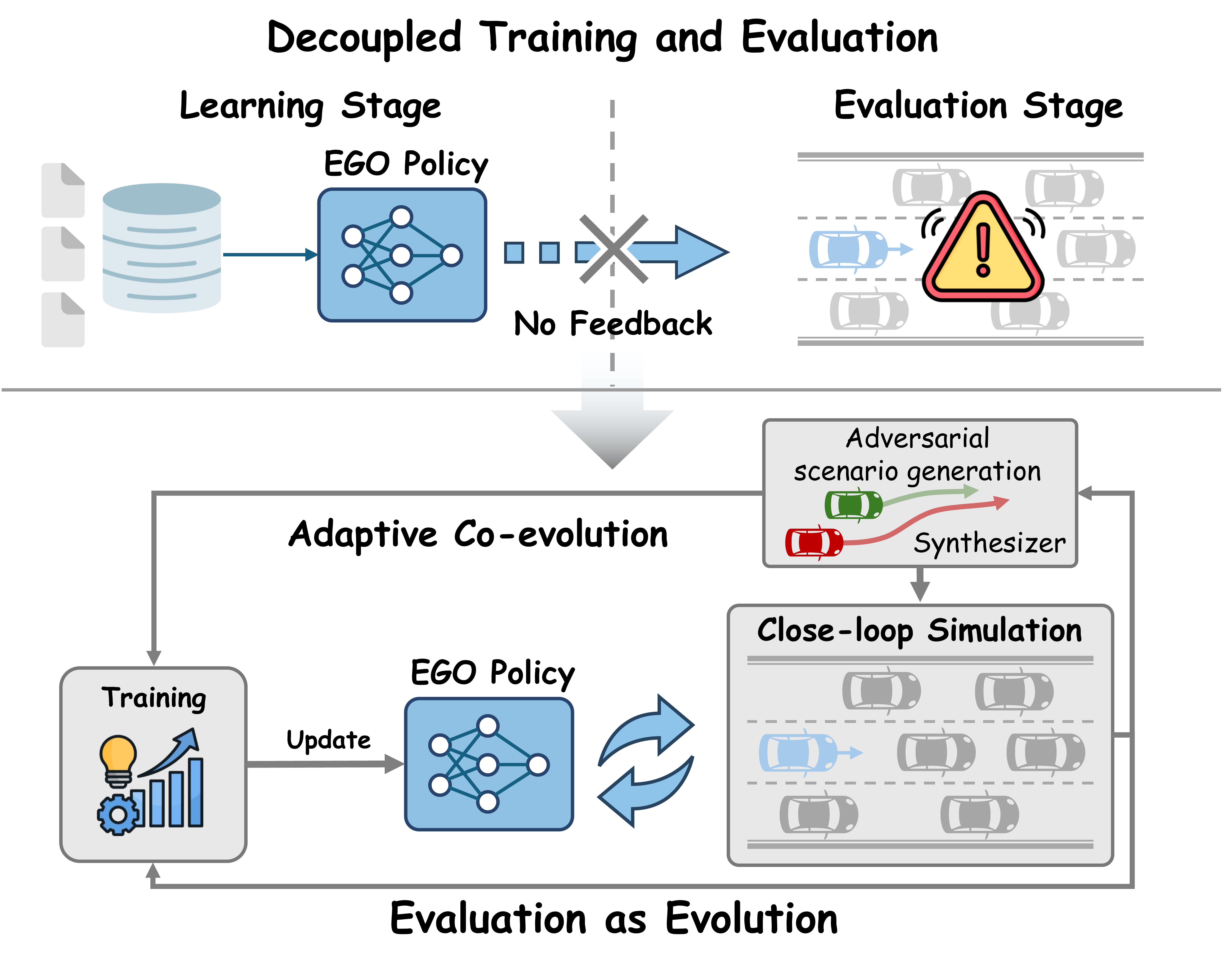}
  \caption{\textbf{From decoupled evaluation to closed-loop evolution.} \emph{Top:} Conventional pipelines separate training from evaluation, so discovered failures remain disconnected from policy learning. \emph{Bottom:} $E^2$ couples an Adversarial Synthesizer with an Ego Policy to synthesize adversarial interactions for closed-loop simulation and recycle outcomes as learning signals to update the ego policy.}
  \label{fig:introduction}
\end{figure}

This mismatch exposes a limitation in current development pipelines. Evaluation is frequently treated as a static, post-hoc phase focused on aggregate metrics and a handful of failure examples~\cite{jia2024bench2drive, xiao2025spatialtree}. Even when critical counterexamples are discovered, they are often logged as incident reports or manually designed test cases rather than being systematically converted into training signals~\cite{yang2025embodiedbench}. As a result, the computationally expensive process of discovering boundary failures, which is often the most informative part of development, does not reliably trigger updates that reshape policy behavior~\cite{biagiola2024boundary}. In autonomous driving scenarios where failure modes evolve alongside the agent, an evaluation process that fails to close the loop is inefficient.

We therefore propose a shift in perspective: \textbf{Evaluation as Evolution}, illustrated in Fig.~\ref{fig:introduction}.
In this framework, adversarial generation is not merely a stress test but serves as a closed-loop evolutionary curriculum. An Adversarial Synthesizer actively constructs interactions to expose current weaknesses, and the Ego Policy evolves by optimizing on these experiences. Discovered failures are thus treated as high-value learning signals that localize the decision boundary within the space of multi-agent trajectories.

To achieve this, we formulate adversarial scene synthesis as transport-regularized hybrid optimal control over reverse-time SDEs~\cite{cui2025elucidating}.
A learned SDE prior captures nominal, realistic multi-agent motion, while a control input steers samples toward safety-critical failures, subject to a transport cost that enforces behavioral realism.
Since controlling the full joint scene is intractable, we first perform dimensionality reduction via topological bifurcation analysis. This identifies a sparse subset of agents whose interventions can induce qualitatively distinct interactions. We then apply a semantic feasibility operator to filter physically implausible candidates, yielding a coarse trajectory proposal that preserves the intended interaction structure.

We further introduce Topological Anchoring to stabilize the generation process. Instead of initializing reverse-time sampling purely from noise, we inject the coarse proposal at an intermediate timestep, imposing a boundary condition that preserves the interaction logic while retaining stochastic flexibility. Conditioned on this initialization, we optimize a time-dependent drift controller restricted to the sparse support set, a technique we term Structure-Aware Sparse Control, to steer trajectories toward the failure set under transport regularization. Overall, $E^2$ implements this evolutionary curriculum as tractable control over an SDE trajectory prior, generating adversarial scenarios that expose failures without sacrificing fidelity to the realistic data distribution.

The main contributions are summarized as follows:
\sloppy
\begin{itemize}
    \item We introduce \textbf{Evaluation as Evolution}, a closed-loop framework that transforms adversarial testing from a static validation step into an adaptive evolutionary curriculum. By coupling the adversarial synthesizer directly with policy learning, $E^2$ continuously converts discovered boundary failures into corrective supervision, thereby actively shaping the ego agent's robustness.
    \item We formulate adversarial synthesis as transport-regularized sparse control over a reverse-time SDE prior. To make this high-dimensional optimization tractable, we integrate Topological Bifurcation Analysis for efficient support selection and propose Topological Anchoring to stabilize initialization. This hybrid approach enables targeted failure discovery while strictly constraining generated behaviors to the realistic data distribution.
    \item We validate $E^2$ through closed-loop simulations on nuScenes and nuPlan. It improves failure discovery by 9.01\% on nuScenes and achieves a 21.43\% zero-shot gain on nuPlan over state-of-the-art baselines, all while maintaining realism. More importantly, fine-tuning on these generated boundary cases significantly improves downstream policy robustness.
\end{itemize}

\begin{figure*}
    \centering
    \includegraphics[width=\linewidth]{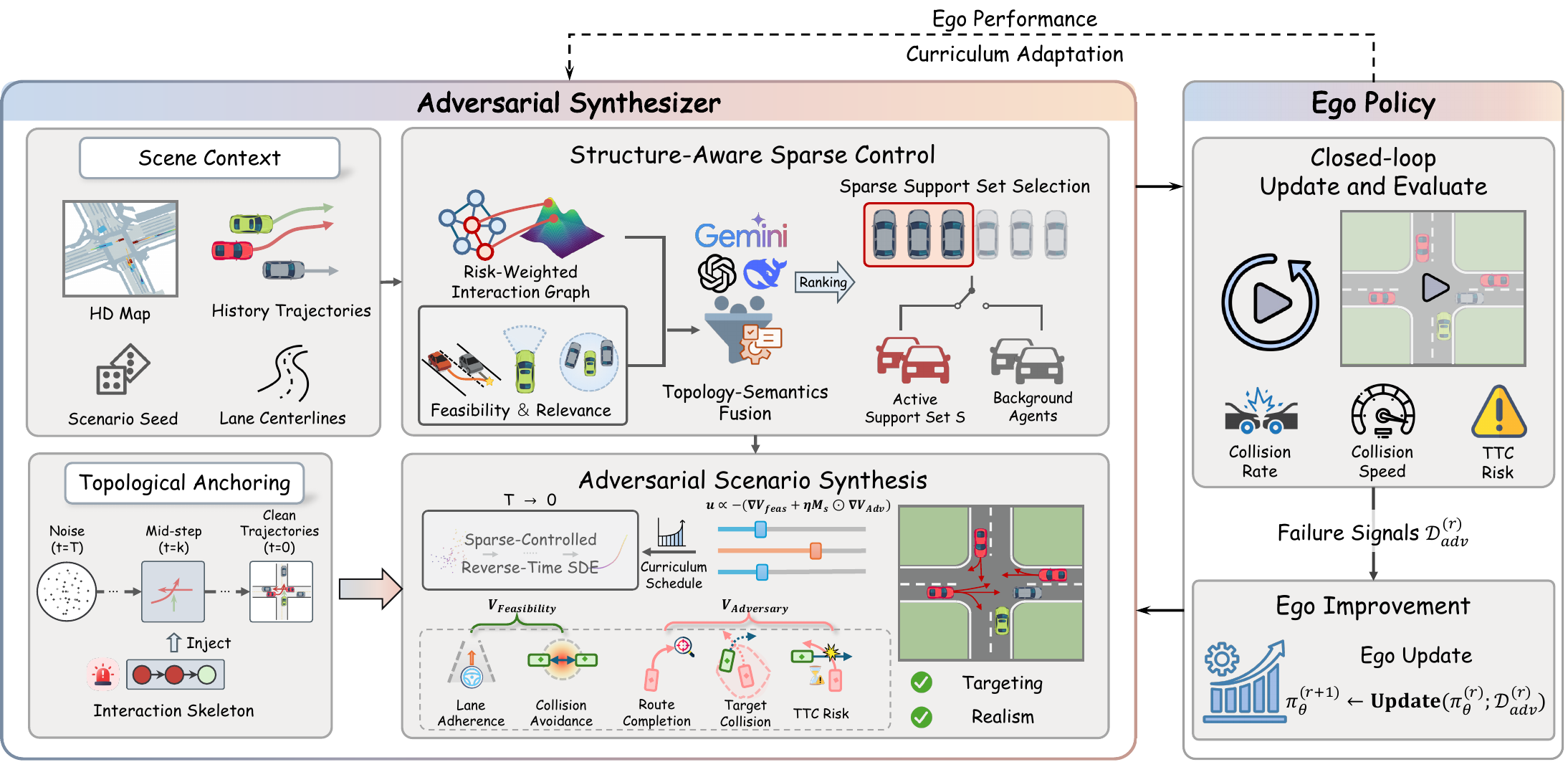}
    \caption{\textbf{Overview of Evaluation as Evolution.} \emph{Left (Adversarial Synthesizer):} given scene context, the Synthesizer builds a risk-weighted interaction graph, selects an intervention-critical Top-$K$ adversary set via topological bifurcation analysis, and synthesizes feasible, realistic adversarial trajectories by transport-regularized sparse control over a reverse-time SDE prior with Topological Anchoring. \emph{Right (Ego Policy):} the Ego Policy executes closed-loop simulation to obtain safety signals for updating the ego policy; the updated performance feeds back to adapt the Synthesizer’s curriculum.}
    \label{fig:framework}
\end{figure*}

\section{Related Work}
\label{sec:related_work}

\subsection{Adversarial Testing and Rare-Event Simulation}
\label{sec:rw_res}

Existing approaches to safety evaluation primarily focus on optimization-based falsification and rare-event simulation~\cite{ljungbergh2024neuroncap,long2025survey,guo2025interactive}. Falsification methods typically search across scenario parameters, initial conditions, or policy spaces to identify violations of safety specifications. 
However, these techniques often scale poorly to high-dimensional multi-agent trajectory spaces and frequently rely on hand-crafted, low-dimensional control variables~\cite{kuznietsov2024explainable, liu2025generating}. 
Rare-event simulation alternatively targets tail events through importance sampling or adaptive cross-entropy methods, but it commonly assumes simplified stochastic models or coarse latent variables, which limits interaction-level objectives and can compromise the realism of the simulation~\cite{ljungbergh2024neuroncap,long2025survey,guo2025interactive}. Distinct from these approaches, our framework conducts adversarial search directly within the trajectory space, employing a KL divergence regularizer to strictly manage deviations from the nominal distribution.

\subsection{Generative Traffic Simulation and Controllable Diffusion}
\label{sec:rw_diffusion}

Generative models serve as powerful data-driven priors for modeling multi-agent futures and facilitating inference-time steering~\cite{liu2024ddm, liao2025diffusiondrive, wang2025generative,zeng2025futuresightdrive}. In particular, diffusion models have emerged as a promising tool for traffic generation and adversarial scenario synthesis, enabling controllable generation through guidance terms derived from constraint potentials or differentiable cost functions~\cite{xu2025survey,pronovost2023scenario,ran2024towards,rowe2025scenario}. A critical challenge, however, lies in maintaining realism during intervention: overly strong guidance often yields implausible behaviors, while heuristic selection of control targets lacks a principled metric for quantifying distribution shift~\cite{li2024drivingdiffusion,cao2024survey}. We mitigate these limitations by formulating guidance as a transport-regularized optimal control problem and by allocating control through structure-aware sparsity.

\subsection{Closed-Loop Policy Learning for Autonomous Driving}
\label{sec:rw_curricula}

Techniques such as curriculum learning and self-play enhance agent robustness by adaptively structuring the training task distribution~\cite{zhu2024advancing,guo2024mappo,xia2025agent0}, while recent advances in autonomous driving emphasize the importance of continual challenge creation through environment-agent co-evolution~\cite{gao2025survey, meng2025preserving,liu2025agent0}. Building on these concepts, researchers in autonomous driving increasingly advocate for closing the loop between evaluation and training. In practice, however, stress testing is frequently treated as a static, post-hoc phase, and discovered failures are rarely recycled into systematic training curricula~\cite{fu2025orion, liu2025reinforced,xing2025openemma}. We instantiate this closed-loop principle by positioning adversarial generation not merely as a testing mechanism, but as a realism-regularized synthesizer that continuously produces an evolving, corrective curriculum for the agent.

\section{Preliminaries}
\label{sec:prelim}
We model the interactive environment as a stochastic generator \(\mathcal{G}_\psi\) that samples multi-agent trajectories. Let the joint state of \(N\) agents at time \(t\) be \(x_t \in \mathbb{R}^{Nd}\), defining a continuous-time trajectory over horizon \(T\) as \(\tau := \{x_t\}_{t \in [0, T]}\). A nominal traffic model induces a path measure \(p_0(\tau)\), with marginal \(p_0(x_0)\), that concentrates on plausible interactions. Adversarial evaluation targets rare failures by biasing the sampling distribution away from \(p_0\), while explicitly regularizing this deviation via a transport-based (KL) penalty. In our framework, we implement this bias as drift control within a reverse-time generative SDE.

\subsection{Reverse-Time SDE Prior for Traffic Dynamics}
\label{sec:sde_prelim}

We use a score-based diffusion model as a nominal prior over multi-agent traffic trajectories. Let \(p_t(\cdot)\) be the marginal density and \(s_\phi(x,t)\approx \nabla_x \log p_t(x)\) the learned score. The corresponding reverse-time SDE is:
\begin{equation}
\mathrm{d}x_t
=
\Big(f(x_t,t) - g(t)^2\, s_\phi(x_t,t)\Big)\,\mathrm{d}t
+
g(t)\,\mathrm{d}\bar{w}_t
\label{eq:reverse_sde}
\end{equation}
which we integrate from \(t=T\) to \(t=0\) with \(x_T \sim \mathcal{N}(0,I)\).
Here \(f\) and \(g\) specify the drift and diffusion schedule, and \(\bar{w}_t\) is a standard Wiener process. The score term encodes data-driven interaction structure (multi-agent couplings and constraints), guiding reverse-time samples toward the realistic data distribution and producing credible traffic rollouts.

\subsection{Adversarial Generation via Drift Control}
\label{sec:control_prelim}

To synthesize adversarial trajectories, we augment the reverse-time dynamics with a drift control \(u_t \in \mathbb{R}^{Nd}\):
\begin{equation}
\mathrm{d}x_t
=
\Big(f(x_t,t) - g(t)^2\, s_\phi(x_t,t) + u_t\Big)\,\mathrm{d}t
+
g(t)\,\mathrm{d}\bar{w}_t
\label{eq:controlled_reverse_sde}
\end{equation}
which induces a controlled path measure \(p_u(\tau)\) when integrated from \(t=T\) to \(t=0\) with \(x_T \sim \mathcal{N}(0,I)\).

We optimize \(u\) by minimizing a realism-regularized adversarial objective:
\begin{small}
\begin{equation}
J(u)
=
\mathbb{E}_{\tau \sim p_u}
\left[
\Phi(x_0)
+
\int_{0}^{T}
\Big(\ell(x_t,t) + \lambda\,\mathcal{R}(u_t,t)\Big)\,\mathrm{d}t
\right]
\label{eq:objective}
\end{equation}
\end{small}
where \(\Phi(x_0)\) encodes the terminal failure or risk criterion, \(\ell\) is an optional running term, and \(\mathcal{R}\) penalizes deviations from the nominal prior.

A standard choice to preserve realism is the quadratic control energy:
\begin{equation}
\mathcal{R}(u_t,t) = \tfrac{1}{2}\,\big\|g(t)^{-1}u_t\big\|_2^2
\label{eq:quad_cost}
\end{equation}
which yields the path-space identity:
\begin{small}
\begin{equation}
\mathrm{KL}\!\left(p_u(\tau)\,\|\,p_0(\tau)\right)
=
\mathbb{E}_{\tau \sim p_u}
\left[
\int_{0}^{T}
\tfrac{1}{2}\,\big\|g(t)^{-1}u_t\big\|_2^2\,\mathrm{d}t
\right]
\label{eq:girsanov_kl}
\end{equation}
\end{small}
Thus, the regularizer provides explicit control over distribution shift from \(p_0\), enabling \(u_t\) to concentrate probability mass on rare failures while preserving behavioral realism.

\section{Methodology}
\label{sec:method}

In this section, we formulate adversarial evaluation as optimal control over the reverse-time SDE. The complete procedure of our proposed $E^2$ framework is summarized in Algorithm \ref{alg:e2}. We first identify a sparse subset of critical agents to ensure computational tractability (Sec.~\ref{sec:sparse_prior}). Then, we implement a hybrid control scheme to stabilize the generation process and preserve behavioral realism (Sec.~\ref{sec:anchoring}). Finally, we integrate this synthesizer into a closed-loop curriculum to iteratively refine the ego policy (Sec.~\ref{sec:closed_loop}).

\begin{algorithm}[t]
\caption{$E^2$: Evaluation as Evolution}
\label{alg:e2}
\begin{algorithmic}[1]
\renewcommand{\algorithmicrequire}{\textbf{Input:}}
\renewcommand{\algorithmicensure}{\textbf{Output:}}
\REQUIRE Dataset $\mathcal{D}$; simulator $\mathrm{Sim}$; reverse-time SDE prior $(f, g, s_\phi)$; initial ego policy $\pi_{\theta^{(0)}}$; rounds $R$; intensity schedule $\{\eta^{(r)}\}$; anchoring parameters $(\alpha, t_a)$.
\ENSURE Updated ego policy $\pi_{\theta^{(R)}}$.

\FOR{$r = 0, \dots, R-1$}
    \STATE Sample a batch of scenes $\mathcal{B} \subset \mathcal{D}$ and initialize buffer $\mathcal{D}'^{(r)} \leftarrow \emptyset$
    \FOR{each scene $s \in \mathcal{B}$}
        \STATE $\tau^{\mathrm{ref}} \leftarrow \mathrm{Sim}(s, \pi_{\theta^{(r)}}, \text{prior})$ 
        \STATE $\mathcal{S}_{\mathrm{top}} \leftarrow \text{TopologicalBifurcation}(\tau^{\mathrm{ref}})$
        \STATE $\mathcal{F} \leftarrow \text{SemanticFeasibilityReasoning}(\mathcal{S}_{\mathrm{top}}, s)$
        \STATE $\mathcal{S} \leftarrow \mathcal{S}_{\mathrm{top}} \cap \mathcal{F}$
        \STATE $(\tilde{\pi}, M_{\mathcal{S}}) \leftarrow \text{InteractionSkeleton}(\mathcal{S}, s)$
        \STATE Initialize latent state $x_T \sim \mathcal{N}(0, I)$
        
        \FOR{$k = K, \dots, 1$}
            \STATE $\hat{x}_0 \leftarrow \hat{x}_0(x_{t_k}, t_k)$ 
            \STATE Compute guided drift control via masked gradients:
            \STATE $\begin{aligned}
                u_{t_k} \leftarrow -g(t_k)^2 \Big[ & \nabla_{x_{t_k}} V_{\mathrm{feas}}(\hat{x}_0) \\
                & + \eta^{(r)} M_{\mathcal{S}} \nabla_{x_{t_k}} V_{\mathrm{adv}}(\hat{x}_0; \tilde{\pi}) \Big]
            \end{aligned}$
            
            \IF{$t_k = t_a$}
                \STATE Apply Topological Anchoring:
                \STATE $x_{t_a} \leftarrow (1-\alpha) x_{t_a} + \alpha \tilde{x}_{t_a}(\tilde{\pi})$
            \ENDIF
            \STATE $x_{t_{k-1}} \leftarrow \mathrm{EulerMaruyama}(x_{t_k}; f, g, s_\phi, u_{t_k})$
        \ENDFOR
        \STATE $\tau^{\mathrm{adv}} \leftarrow \mathrm{Sim}(s, \pi_{\theta^{(r)}}, \text{env}(x_0))$ 
        \STATE $\mathcal{D}'^{(r)} \leftarrow \mathcal{D}'^{(r)} \cup \{(s, \tau^{\mathrm{adv}})\}$
    \ENDFOR
    \STATE \textbf{Policy fine-tuning:} 
    \STATE $\theta^{(r+1)} \leftarrow \mathrm{Update}(\theta^{(r)}; \mathcal{D}'^{(r)})$
\ENDFOR
\STATE \textbf{return} $\pi_{\theta^{(R)}}$
\end{algorithmic}
\end{algorithm}

\subsection{Structure-Aware Sparse Control}
\label{sec:sparse_prior}
Applying adversarial control to all agents in dense traffic is computationally prohibitive and often disrupts background realism. We address this by restricting the control drift to a sparse subset of critical agents. Let \(\mathcal{I}=\{1,\dots,N\}\) denote the agent index set. We implement this sparsity by applying a deterministic mask to the unrestricted latent control \(v_t\in\mathbb{R}^{Nd}\), thereby isolating the active support set \(\mathcal{S}\subset\mathcal{I}\):
\begin{equation}
u_t \;=\; M_{\mathcal{S}}\, v_t
\label{eq:support_mask}
\end{equation}
where \(M_{\mathcal{S}} := \mathrm{diag}(m_1 I_d,\dots,m_N I_d)\) is a block-diagonal matrix with \(m_i=\mathbf{1}[i\in\mathcal{S}]\). Under a quadratic transport cost, the KL penalty depends only on the active components. This sparsity reduces the effective control dimension and limits the deviation from the nominal measure \(p_0\). To define the active support set \(\mathcal{S}\), we first isolate topologically significant candidates, and subsequently refine them against semantic constraints.

\begin{figure*}[t]
    \centering
    \includegraphics[width=0.9\linewidth]{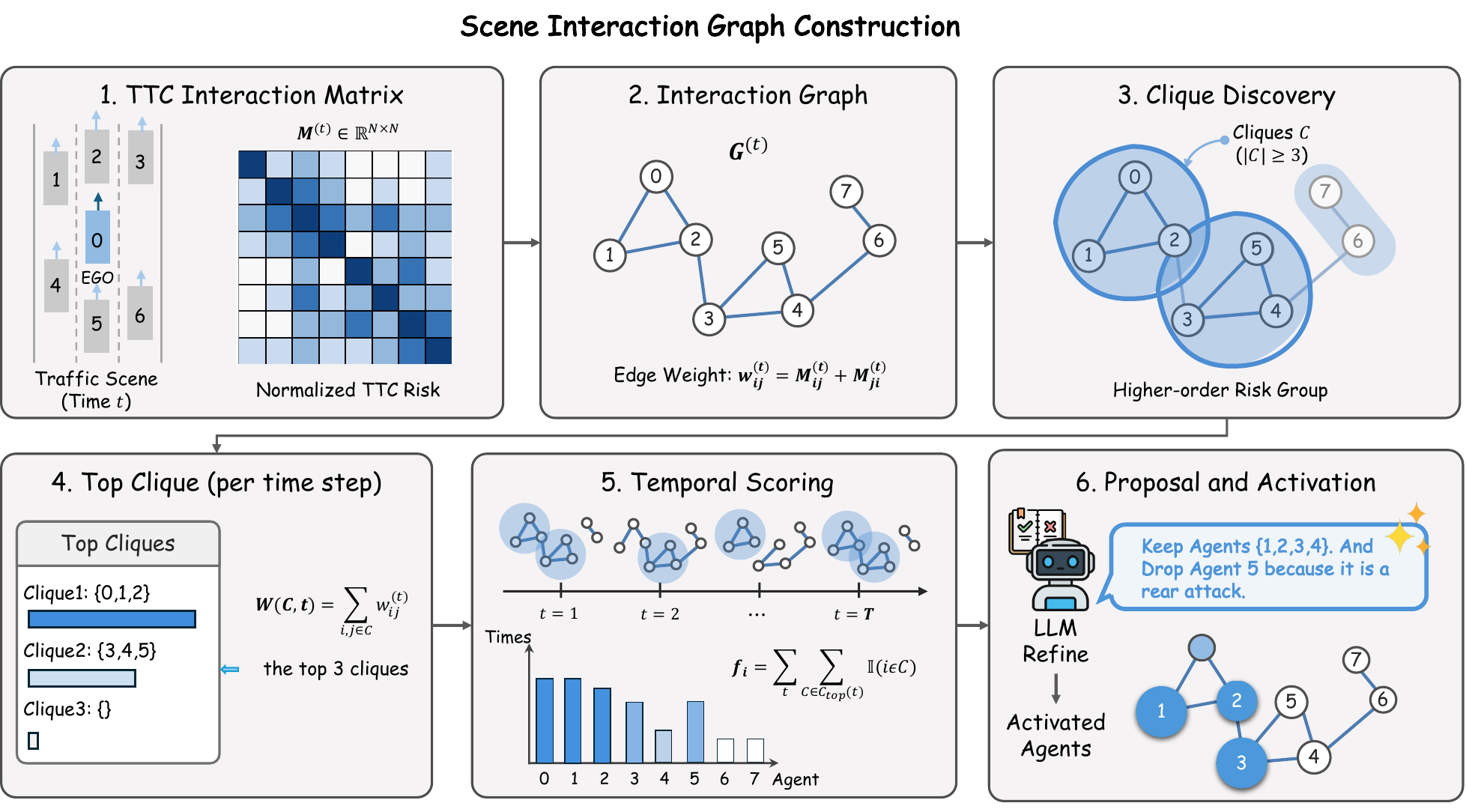}
    \caption{\textbf{Structure-aware sparse control via scene interaction graph construction.}
    The Synthesizer builds a TTC-based risk interaction matrix, converts it into a weighted interaction graph, discovers higher-order
    risk groups (cliques), and selects top cliques across time to score and activate a small subset of adversarial agents for
    targeted control.}
    \label{fig:Scene_Interaction_Graph_Construction}
\end{figure*}

\subsubsection{Topological Bifurcation Analysis}
\label{sec:bifurcation}

Coupling the search for critical agents directly with the generative control loop is inefficient. We therefore decouple discovery from generation by inferring the interaction topology from a lightweight nominal rollout.

\textbf{Risk-weighted interaction graph.}
Given a reference rollout \(x^{\mathrm{ref}}_{0:T}:=\{x_t^{\mathrm{ref}}\}_{t\in[0,T]}\), we evaluate the interaction risk between any pair of agents at each timestep. We construct a sequence of undirected interaction graphs \(G^{(t)}=(\mathcal{I},\mathcal{E}^{(t)},W^{(t)})\) for each time \(t \in [0, T]\). Here, \(\mathcal{I}\) indexes the \(N\) agents, and \(W^{(t)}\) assigns an instantaneous risk weight to each pair. For any agents \(i \neq j\), the weight \(w_{ij}^{(t)}\) is derived from a time-to-collision (TTC) surrogate:
\begin{equation}
w_{ij}^{(t)}
\;:=\;
\sigma\!\left(\frac{\tau_{\max}-\mathrm{TTC}_{ij}(x_t^{\mathrm{ref}})}{\beta}\right)
\in[0,1]
\label{eq:ttc_weight_t}
\end{equation}
where \(\sigma(z):=(1+\exp(-z))^{-1}\), \(\tau_{\max}>0\) caps the TTC horizon, and \(\beta>0\) controls the softness. We retain edges above a threshold \(\epsilon_w\in(0,1)\) to filter out trivial pairs:
\begin{equation}
(i,j)\in\mathcal{E}^{(t)}\quad \Longleftrightarrow \quad w_{ij}^{(t)}\ge \epsilon_w
\end{equation}
yielding a sparse graph \(G^{(t)}\) that captures potentially coupled interactions at the exact moment \(t\).

\textbf{Bifurcation points via temporal clique scoring.}
To identify structurally critical interaction nexus, we evaluate the topological properties of \(G^{(t)}\) independently at each timestep. We identify high-risk \(K\)-cliques by computing the intra-clique risk weight \(W(C, t) = \sum_{i,j \in C} w_{ij}^{(t)}\). We select the top-ranked cliques at each step to form the instantaneous critical set \(\mathcal{C}_{\mathrm{top}}^{(t)}\). 

To quantify an agent's structural importance across the entire interaction horizon, we compute a temporal score \(f_i\) that accumulates its occurrences within these critical cliques over time:
\begin{equation}
    f_i = \sum_{t=1}^{T} \sum_{C \in \mathcal{C}_{\mathrm{top}}^{(t)}} \mathbb{I}(i \in C)
    \label{eq:temporal_scoring}
\end{equation}
where \(\mathbb{I}(\cdot)\) is the indicator function. A large \(f_i\) indicates that an agent persistently occupies a structurally sensitive interaction nexus throughout the rollout. Finally, we select the agents with the highest temporal scores \(f_i\) to form the initial topological candidate set \(\mathcal{S}_{\mathrm{top}}\).

\subsubsection{Semantic Feasibility Reasoning}
\label{sec:semantic_filter}

While topological analysis identifies structurally critical candidates, it ignores inherent physical constraints and environmental context. To refine $\mathcal{S}_{\mathrm{top}}$, we apply a semantic feasibility projection $\Pi_{\mathrm{sem}}$ based on an ego conditioned scene description $\mathsf{desc}$ that encodes map topology, lane graphs, and local geometry.

We establish a binary feasibility mask $m_i^{\mathrm{sem}} = \Pi_{\mathrm{sem}}(i;\mathsf{desc}) \in \{0,1\}$, yielding the feasible subset $\mathcal{F} = \{i : m_i^{\mathrm{sem}}=1\}$. The final sparse control support is then determined by the intersection $\mathcal{S} = \mathcal{S}_{\mathrm{top}} \cap \mathcal{F}$.

To ensure deterministic and physically sound interventions, the projection $\Pi_{\mathrm{sem}}$ enforces the following semantic criteria:

\begin{itemize}
    \item \textbf{Physical and map consistency:} Candidates must be located within valid drivable regions and maintain clear lane associations.
    \item \textbf{Interaction potential:} Targeted agents must exhibit a plausible near term route conflict or centerline overlap with the ego vehicle within the planning horizon.
    \item \textbf{Dynamic relevance:} Static or parked vehicles lacking a direct blocking effect are rejected, thereby focusing control strictly on agents capable of active and meaningful interaction.
\end{itemize}

This logic is implemented via a lightweight language model that evaluates scene cues, such as proximity to intersections or merges, to confirm whether an agent is a semantically viable adversarial target. By grounding the support set in these traffic-specific priors, the synthesizer effectively isolates kinematically valid and dynamically relevant agents for targeted control.

\subsubsection{Interaction Skeleton Construction}
\label{sec:interaction_skeleton}

Given the final sparse control support $\mathcal{S}$, we deterministically instantiate an interaction skeleton $\tilde{\pi}$ via Algorithm \ref{alg:skeleton}. This skeleton acts as a structural prior by extracting an active coalition $C^\star \subseteq \mathcal{S}$ and assigning a causal dependency structure.

We partition $\mathcal{S}$ into proximal interactors $C_{\text{near}}$ and distal initiators $C_{\text{far}}$, establishing $C^\star = C_{\text{near}} \cup C_{\text{far}}$. Proximal agents form the final link in the failure pathway and are selected by minimizing their aggregate Euclidean distance $d(i, e)$ to the ego $e$:
\begin{equation}
    C_{\text{near}} = \underset{\substack{C \subset \mathcal{S} \\ |C| = K_{\text{near}}}}{\operatorname{arg\,min}} \sum_{i \in C} d(i, e)
\end{equation}

Distal initiators influence the scene dynamics from a distance. We first isolate a candidate pool $\mathcal{S}_{\text{far}}$ by excluding proximal agents, enforcing a minimum distance threshold $\tau_{\text{dist}}$, and requiring non-adjacency to the ego's lane $\mathcal{L}_{\text{adj}}(e)$:
{\small
\begin{equation}
    \mathcal{S}_{\text{far}} = \left\{ i \in \mathcal{S} \setminus C_{\text{near}} \mid d(i, e) > \tau_{\text{dist}} \text{ and } i \notin \mathcal{L}_{\text{adj}}(e) \right\}
\end{equation}
}

From $\mathcal{S}_{\text{far}}$, we select the $K_{\text{far}}$ furthest agents to maximize long-range interaction potential:
\begin{equation}
    C_{\text{far}} = \underset{\substack{C \subset \mathcal{S}_{\text{far}} \\ |C| = K_{\text{far}}}}{\operatorname{arg\,max}} \sum_{i \in C} d(i, e)
\end{equation}

To assign causal ordering, we find a kinematically viable, transitive influence pathway from an initiator $f \in C_{\text{far}}$ to a target $n \in C_{\text{near}}$, constrained by a longitudinal headway $\tau_{\text{long}}$. The primitive $\mathcal{P}$ then instantiates $\tilde{\pi}$ by reconfiguring the adversarial targets to enforce a cascading dependency. 
Instead of all agents targeting the ego simultaneously, $f$ engages an intermediate agent, which subsequently targets $n$, and only $n$ directly engages the ego. This gradient chaining efficiently propagates adversarial coupling.

Because $|C^\star| \ll N$, computing this skeleton adds negligible overhead to the inference loop. If no valid intermediate agents exist, the coalition robustly defaults to $C_{\text{near}}$ for direct local interaction, and kinematically infeasible configurations are pruned. 

Ultimately, $\tilde{\pi}$ determines the anchoring configuration $\tilde{x}_{t_a}$. We first instantiate the causal graph $\tilde{\pi}$ into a continuous, kinematically feasible joint configuration $\tilde{x}_0$ via a coarse lane-graph planner. We then project this clean state into the intermediate diffusion timestep $t_a$ according to the forward diffusion process~\cite{ho2020denoising}:
\begin{equation}
    \tilde{x}_{t_a} = \sqrt{\bar{\gamma}_{t_a}} \tilde{x}_0 + \sqrt{1 - \bar{\gamma}_{t_a}} \epsilon, \quad \epsilon \sim \mathcal{N}(0, I)
\end{equation}
where $\bar{\gamma}_{t_a} = \prod_{j=0}^{t_a} \gamma_j$ represents the cumulative product of the incremental noise reduction terms $\gamma_j := 1 - \beta_j$ up to $t_a$, with $\beta_j$ dictated by the predefined noise schedule. This formulation guarantees that the anchored state remains consistent with the required noise distribution.

\subsection{Hybrid Optimal Control via Topological Anchoring}
\label{sec:anchoring}
Given the identified support \(\mathcal{S}\) and skeleton \(\tilde{\pi}\), we formulate adversarial generation as optimal control over the reverse-time SDE. To ensure stability and maintain fidelity to the nominal prior, we propose a hybrid scheme coupling topological anchoring with sparse gradient guidance.

\subsubsection{Topological Anchoring}
\label{sec:topo_anchor}

Rare coordinated failures occupy low-probability regions under the nominal prior \(p_0(\tau)\). Consequently, reverse-time integration from noise \(x_T\sim\mathcal{N}(0,I)\) is often unstable, as the early denoising steps must reconstruct a plausible scene while selecting a rare interaction mode. We stabilize generation by anchoring at an intermediate time \(t_a\in(0,T)\) using the interaction skeleton \(\tilde{\pi}\). Specifically, after integrating Eq.~\eqref{eq:reverse_sde} from \(T\) to \(t_a\), we blend the state with a skeleton-consistent configuration \(\tilde{x}_{t_a}\):
\begin{equation}
x_{t_a}
\;\leftarrow\;
(1-\alpha)\,x_{t_a} + \alpha\,\tilde{x}_{t_a}
\qquad \alpha\in(0,1]
\label{eq:anchor_injection}
\end{equation}

This blending operation acts as a structural soft constraint at the intermediate step, pulling the reverse time sampling distribution toward a valid interaction mode prior to the application of sparse gradient based control.

\subsubsection{Sparse Gradient Control}
\label{sec:sparse_control}

We then implement KL-regularized drift control as score-based gradient guidance on the denoised prediction \(\hat{x}_0(x_t,t)\). This reduces the path objective to a potential defined on \(\hat{x}_0\), decomposed into feasibility and targeted adversarial terms:
\begin{equation}
\begin{aligned}[b]
\Phi(x_0)+\int_0^T \ell(x_t,t)\,dt
\;&\approx\;
V_{\mathrm{feas}}(\hat{x}_0)
+\eta\,V_{\mathrm{adv}}(\hat{x}_0;\tilde{\pi},\mathcal{S})
\end{aligned}
\label{eq:potential_decomp}
\end{equation}
where \(V_{\mathrm{feas}}\) enforces physical and traffic-rule constraints and \(V_{\mathrm{adv}}\) promotes the failure pattern \(\tilde{\pi}\) on the active set \(\mathcal{S}\). The gain \(\eta\ge 0\) sets the risk sensitivity.

To limit the deviation from the prior, we apply feasibility gradients to all agents but localize adversarial gradients to \(\mathcal{S}\):
\begin{small}
\begin{equation}
u_t=
-g(t)^2\!\left(\nabla_{x_t} V_{\mathrm{feas}}(\hat{x}_0)
+\eta\,M_{\mathcal{S}}\nabla_{x_t} V_{\mathrm{adv}}(\hat{x}_0;\tilde{\pi},\mathcal{S})\right)
\label{eq:masked_guidance}
\end{equation}
\end{small}
so agents outside \(\mathcal{S}\) follow the nominal score-induced reverse dynamics up to feasibility corrections, preserving background realism while focusing adversarial control on the selected agents.

We discretize the controlled reverse-time SDE with the Euler-Maruyama method on \(\{t_k\}_{k=0}^K\) (with \(\Delta t_k=t_{k-1}-t_k<0\)):
\begin{small}
\begin{equation}
\begin{aligned}[b]
x_{t_{k-1}}
&= x_{t_k}
+ \Big(
    f(x_{t_k},t_k)
    - g(t_k)^2 s_\phi(x_{t_k},t_k)
    + u_{t_k}
  \Big)\Delta t_k \\
&\quad + g(t_k)\sqrt{|\Delta t_k|}\,\xi_k,
\qquad \xi_k\sim\mathcal{N}(0,I)
\label{eq:Euler-Maruyama}
\end{aligned}
\end{equation}
\end{small}
and inject topological anchoring once at \(t_{k_a}\approx t_a\) via Eq.~\eqref{eq:anchor_injection}. Together, anchoring and sparse masking concentrate samples near the critical set while keeping non-interacting agents consistent with the SDE prior.

\begin{algorithm}[h]
   \caption{Interaction Skeleton Construction}
   \label{alg:skeleton}
\begin{algorithmic}[1]
   \renewcommand{\algorithmicrequire}{\textbf{Input:}}
   \renewcommand{\algorithmicensure}{\textbf{Output:}}
   \REQUIRE Filtered support set $\mathcal{S}$, Ego state $e$, structural thresholds $(\tau_{\text{dist}}, \tau_{\text{long}})$, capacities $(K_{\text{near}}, K_{\text{far}})$.
   \ENSURE Interaction skeleton $\tilde{\pi}$, binary mask $M_{\mathcal{S}}$.
   \STATE $C_{\text{near}} \leftarrow \underset{C \subset \mathcal{S}, |C| = K_{\text{near}}}{\operatorname{arg\,min}} \sum_{i \in C} d(i, e)$
   \STATE $\mathcal{S}_{\text{far}} \leftarrow \{ i \in \mathcal{S} \setminus C_{\text{near}} \mid d(i, e) > \tau_{\text{dist}} \land i \notin \mathcal{L}_{\text{adj}}(e) \}$
   \STATE $C_{\text{far}} \leftarrow \underset{C \subset \mathcal{S}_{\text{far}}, |C| = K_{\text{far}}}{\operatorname{arg\,max}} \sum_{i \in C} d(i, e)$
   \STATE $C^\star \leftarrow C_{\text{near}} \cup C_{\text{far}}$
   \IF{no kinematically viable chain exists between $C_{\text{far}}$ and $C_{\text{near}}$}
       \STATE $C^\star \leftarrow C_{\text{near}}$ 
   \ENDIF
   \STATE Define binary mask $M_{\mathcal{S}} = \text{diag}(\mathbf{1}[i \in C^\star])$
   \STATE Instantiate $\tilde{\pi}$ by re-targeting $V_{\mathrm{adv}}$ within $C^\star$: $f \to \dots \to n \to e$
   
   \STATE \textbf{return} $(\tilde{\pi}, M_{\mathcal{S}})$
\end{algorithmic}
\end{algorithm}

\subsection{Closed-Loop Evolutionary Curricula}
\label{sec:closed_loop}

We integrate the adversarial synthesizer into a closed-loop curriculum. Instead of relying on a static set of test scenarios, the data generation and policy training advance iteratively. At each round \(r\), the synthesizer analyzes the current policy \(\pi_{\theta_r}\) to construct targeted boundary cases. By updating the support set \(\mathcal{S}_r\) and interaction skeleton \(\tilde{\pi}_r\) based on the policy's recent performance, the control mechanism efficiently samples the active failure frontier.

The ego policy then updates to \(\pi_{\theta_{r+1}}\) using these targeted rollouts, typically via reinforcement learning or constrained optimization. This iterative process creates a continuously shifting data distribution. As the policy improves, previously critical interactions are resolved. The generation framework adapts to this improvement by identifying new bifurcation points, ensuring that the generated scenarios remain informative and challenging.

To regulate the difficulty of this process, we schedule the adversarial intensity \(\eta_r\). Initial rounds use a low \(\eta_r\) to expose basic errors close to the nominal traffic prior. As the policy strengthens, increasing \(\eta_r\) enables the synthesizer to explore larger state deviations, revealing complex failures that require strict multi-agent coordination. This phased approach maintains a consistent learning gradient and facilitates continuous robustness improvements in the high-dimensional traffic environment.

\section{Experiments and Results}
\label{sec:experiments}

In this section, we evaluate $E^2$ in a closed-loop multi-agent simulation, aiming to answer the following questions: (1) How does the adversarial quality of $E^2$ compare against state-of-the-art dense guidance baselines? (2) How effective is each key component of our framework? (3) Can the adversarial evaluation capabilities of $E^2$ generalize to expose the distinct failure modes of diverse ego policies? (4) Is the proposed closed-loop curriculum effective at improving the ego policy's robustness?

\subsection{Experimental Setup}
\label{sec:exp_setup}

\subsubsection{Simulation Environment and Dataset}
All evaluations are conducted in a closed-loop multi-agent traffic simulator~\cite{xu2023bits}. At each replanning step, the simulator generates other-agent trajectories conditioned on the scene context and the current ego state. We initialize scenes from the nuScenes~\cite{caesar2020nuscenes} and nuPlan~\cite{caesar2021nuplan} validation splits, focusing on vehicle-to-vehicle interactions. To demonstrate cross-dataset generalizability, our framework is directly deployed on the nuPlan dataset without any model retraining.
Background traffic is generated by a pretrained diffusion trajectory model that is kept frozen, inducing our nominal reverse-time SDE prior. $E^2$ intervenes only through inference-time drift control during denoising.

\subsubsection{Ego Policies}
We evaluate three ego policies: a rule-based lane-graph (LG) policy~\cite{rempe2022generating}, the Intelligent Driver Model (IDM)~\cite{treiber2000congested}, and the hybrid BITS policy~\cite{xu2023bits}. During scenario generation, each ego is treated as a black-box controller. All policies share the same observation interface and replanning frequency.

\subsubsection{Baselines}
We compare against CTG~\cite{zhong2023guided}, STRIVE~\cite{rempe2022generating}, DiffScene~\cite{xu2025diffscene}, CCDiff~\cite{lin2025causal}, and Safe-Sim-opt, our multi-agent extension of Safe-Sim~\cite{chang2024safe}. These baselines primarily apply dense or global guidance across agents or rely on rule-based interventions, without explicitly leveraging interaction topology for support selection.

\subsubsection{Evaluation Metrics}
We classify evaluation metrics into two primary categories: safety criticality and realism.

\textbf{Safety Criticality and Severity.}
These metrics measure the intensity and consequences of the adversarial interaction.
We report the Collision Failure Rate (CFR)~\cite{tan2023language} and the Minimum Distance-to-Collision (MinDTC) to quantify hazard frequency and proximity.
To assess failure severity and mode, we report the Relative Impact Velocity (RelVel) and the Rear-Impact Rate (RIR)~\cite{lin2025causal}. We also include the TTC cost (TTC-C)~\cite{chang2024safe} and the ego mean acceleration (mAcc) as a proxy for the evasive effort required.
Higher CFR and TTC-C indicate increased hazard frequency, while lower MinDTC reflects spatial proximity to collisions. RelVel quantifies impact severity conditional on a collision, characterizing the nature of the interaction.

\textbf{Realism and Validity.}
These metrics assess whether the generated scenarios remain physically plausible and compliant with traffic rules.
We report the All-Off-Road rate (AOR)~\cite{ding2024realgen}, measuring the fraction of rollouts where any vehicle leaves the drivable area; a high AOR indicates invalid attacks.
Additionally, we report a transport fidelity score (REAL) based on the Wasserstein distance between simulated and real kinematic statistics, normalized such that higher values indicate closer agreement with the real data distribution~\cite{zhong2023guided}.

\begin{table}[t]
  \caption{\textbf{Closed-loop performance compared with baselines.} The metrics quantify the realism and safety criticality of generated interactions. We report the mean over 7 random seeds. Best in each column (per dataset) is \textbf{bold} and second-best is \underline{underlined}. \BlueHL{Blue} denotes $E^2$.}
  \label{tab:quality_adversarial_interactions}
  \centering
  \small
  \setlength{\tabcolsep}{4pt}
  \renewcommand{\arraystretch}{1.2}
  \begin{tabular}{lcccc}
    \toprule
    Method &
    REAL &
    \makecell{CFR (\%)} &
    \makecell{AOR (\%)} &
    \makecell{RelVel ($\mathrm{m/s}$)} \\
    
    \midrule
    \multicolumn{5}{c}{\textbf{nuScenes Dataset}} \\
    \midrule
    CTG          & \best{0.8011} & 1.18            & \second{0.20} & \best{4.07} \\
    STRIVE       & 0.6115        & \second{51.28}  & 3.31          & 7.01 \\
    Safe-Sim-opt & 0.7394        & 37.14           & 0.26          & 17.57 \\
    DiffScene    & 0.7325        & 31.43           & 0.38          & 26.07 \\
    CCDiff       & 0.6965        & 39.90           & 4.11          & 33.05 \\
    \rowcolor{oursblue}
    Ours &
      \second{0.7432} &
      \best{60.29} &
      \best{0.14} &
      \second{4.67} \\
      
    \midrule
    \multicolumn{5}{c}{\textbf{nuPlan Dataset}} \\
    \midrule
    Safe-Sim-opt & \second{0.8789} & \second{57.14}  & \second{2.29} & \second{22.77} \\
    \rowcolor{oursblue}
    Ours &
      \best{0.8984} &
      \best{78.57} &
      \best{1.69} &
      \best{21.81} \\
    \bottomrule
  \end{tabular}
\end{table}

\begin{figure*}
    \centering
    \includegraphics[width=0.95\linewidth]{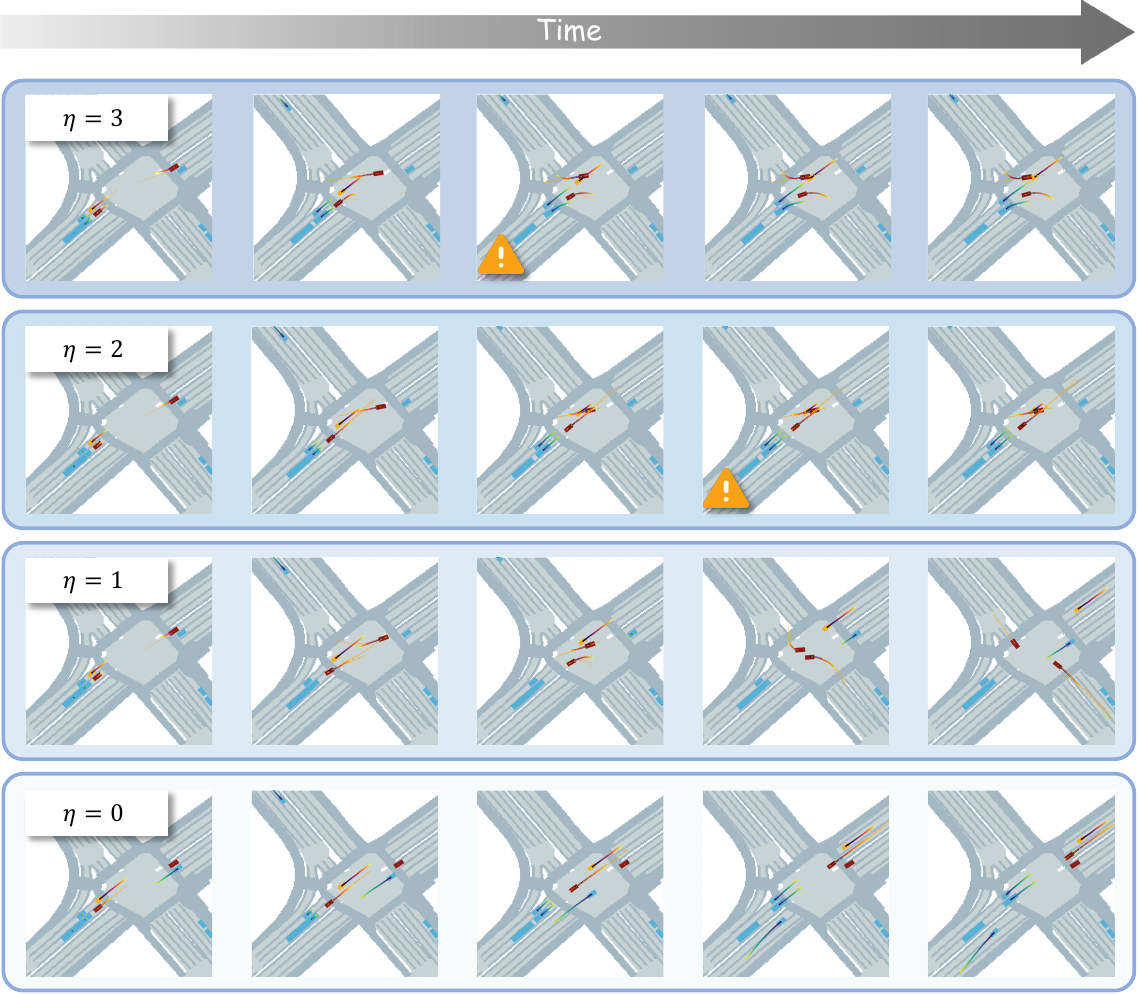}
    \caption{\textbf{Qualitative controllability of failure timing and type under increasing adversarial intensity.}
Closed-loop rollouts for the lane-graph (LG) ego policy under a fixed random seed (seed=50), with $\eta\in\{0,1,2,3\}$ from bottom to top.}
    \label{fig:graph}
\end{figure*}

\begin{table}[t]
  \caption{\textbf{Ablation of structure-aware sparse control under a fixed control budget.} We compare Random-K, Topo-only (bifurcation-based targeting without feasibility filtering), and Topo+Feas. (ours). Results report the mean over 7 random seeds. Best in each column is \textbf{bold} and second-best is \underline{underlined}. \BlueHL{Blue} denotes our method.}
  \label{tab:key_adversary_activation}
  \centering
  \small
  \setlength{\tabcolsep}{6pt}
  \renewcommand{\arraystretch}{1.2}
  \begin{tabular}{lccccc}
    \toprule
    Method &
    \makecell{TTC-C} &
    \makecell{CFR \\ (\%)} &
    \makecell{RIR \\ (\%)} &
    \makecell{AOR \\(\%)} &
    \makecell{RelVel \\ ($\mathrm{m/s}$)} \\
    \midrule
    Random-K & \second{0.0949} & 28.57            & \best{19.99}   & \best{0.15}   & 18.25 \\
    Topo-only    & 0.0925          & \second{31.43}   & 45.47          & 0.32          & \second{15.37} \\
    \rowcolor{oursblue}
    Topo+Feas. &
      \best{0.1800} &
      \best{53.57} &
      \second{26.66} &
      \second{0.20} &
      \best{2.58} \\
    \bottomrule
  \end{tabular}
\end{table}

\subsection{Quality of Adversarial Interactions}
\label{sec:quality_analysis}
\textbf{Quantitative Performance Comparison.} A core challenge in adversarial generation lies in balancing criticality and realism. Table~\ref{tab:quality_adversarial_interactions} demonstrates that $E^2$ effectively navigates this trade-off, achieving the highest failure rate (CFR 60.29\%) while strictly adhering to the valid drivable area (AOR 0.14\%) and maintaining high transport fidelity (REAL 0.7432). By localizing drift control to topology-identified bifurcation agents, our method reaches the failure set via plausible interaction couplings rather than map-inconsistent motions.

In contrast, baseline methods struggle to satisfy these objectives simultaneously. While optimization-based approaches, such as STRIVE, achieve high failure rates (51.28\%), they do so by sacrificing distributional fidelity (REAL 0.6115) and frequently violating map constraints (AOR 3.31\%). Conversely, dense diffusion methods like CTG preserve realism (REAL 0.8011) but fail to discover rare events (CFR 1.18\%), suggesting that global guidance in high-dimensional spaces is inefficient for locating failure modes.
Other baselines report moderate failure rates but incur excessive impact velocities (RelVel $>17~\mathrm{m/s}$). This severity is associated with degraded realism and increased off-road rates, suggesting that these collisions arise from distribution shift and physical violations rather than valid adversarial interactions.

Furthermore, $E^2$ demonstrates exceptional cross-dataset generalizability. As shown in Table~\ref{tab:quality_adversarial_interactions}, when deployed directly on the nuPlan dataset without any model retraining, our method achieves a CFR of 78.57\%, outperforming the baseline by an absolute margin of 21.43\%. Crucially, this significant increase in failure discovery does not compromise validity, as evidenced by the highest realism score and the lowest rate of off-road violations. This confirms that our framework effectively generalizes to unseen environmental configurations and traffic distributions.

\textbf{Controllability of Adversarial Intensity.} Beyond evaluating aggregate metrics, $E^2$ provides precise control over the adversarial intensity through the parameter $\eta$. As illustrated in Fig.~\ref{fig:graph}, smoothly scaling this value directly dictates both the severity and the timing of the generated failures. At lower intensities, the adversary favors subtle and localized perturbations that typically manifest as lateral collisions late in the scenario. Conversely, increasing the adversarial intensity prompts the synthesizer to trigger highly aggressive behaviors among multiple agents, resulting in earlier collisions and severe frontal impacts. This predictable relationship between intensity and failure mode confirms that our framework serves as a reliable mechanism to structure progressive and hierarchical safety curricula.

\begin{table}[t]
  \caption{\textbf{Ablation of drift control components and Topological Anchoring.} We evaluate combinations of the feasibility potential $V_{\mathrm{feas}}$ and the adversarial potential $V_{\mathrm{adv}}$, with anchoring enabled or disabled.}
  \label{tab:guidance_pi_ablation}
  \centering
  \small
  \setlength{\tabcolsep}{5.0pt}
  \renewcommand{\arraystretch}{1.15}

  \begin{tabular*}{\columnwidth}{@{\extracolsep{\fill}} cc ccccc}
    \toprule
    \multicolumn{2}{c}{Drift Control} & \multicolumn{5}{c}{Metrics} \\
    \cmidrule(lr){1-2}\cmidrule(lr){3-7}
    $V_{\mathrm{feas}}$ & $V_{\mathrm{adv}}$ &
    TTC-C & \makecell{CFR\\(\%)} & \makecell{RelVel\\($\mathrm{m/s}$)} & REAL & \makecell{AOR\\(\%)} \\
    \midrule

    \rowcolor{gray!10}\multicolumn{7}{c}{\scshape Anchoring disabled} \\
    \addlinespace[1pt]
    $\checkmark$ & \na &
    0.0722 & 3.57 & 3.67 & 0.7648 & 0.19 \\
    \na & $\checkmark$ &
    0.2103 & 87.86 & 17.01 & 0.7286 & 1.44 \\
    $\checkmark$ & $\checkmark$ &
    0.1532 & 51.43 & 2.11 & 0.7464 & 0.10 \\
    \addlinespace[2pt]

    \rowcolor{gray!8}\multicolumn{7}{c}{\scshape Anchoring enabled} \\
    \addlinespace[1pt]
    $\checkmark$ & $\checkmark$ &
    0.1501 & 57.14 & 6.56 & 0.7848 & 0.31 \\
    \bottomrule
  \end{tabular*}
  \vskip -0.08in
\end{table}

\begin{figure}[t]
  \centering
  \includegraphics[width=\linewidth]{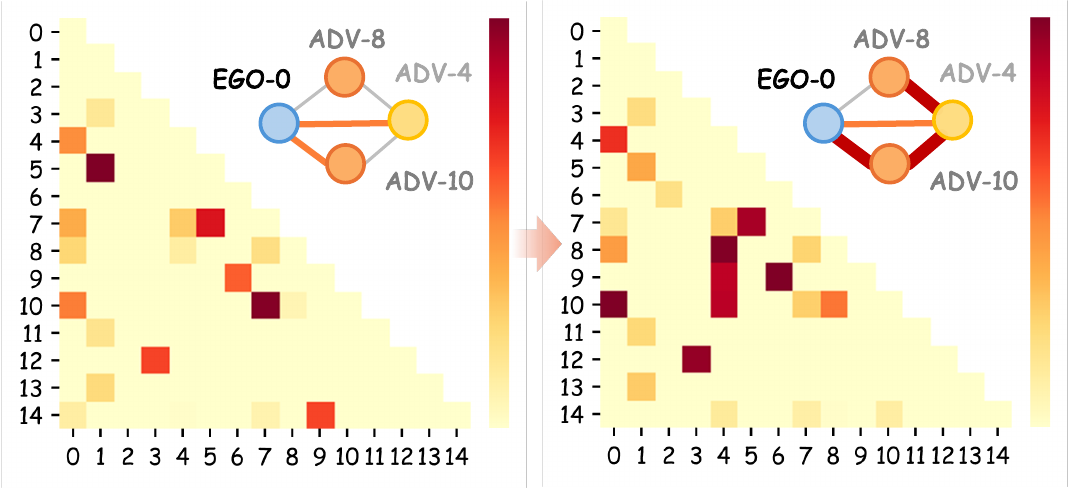}
  \caption{\textbf{Qualitative interaction skeleton.} TTC-based interaction matrices and the activated ego-adversary chain for the same scene. \emph{Left:} Topological Anchoring disabled. \emph{Right:} Topological Anchoring enabled.}
  \label{fig:partial_exp2}
\end{figure}

\subsection{Ablation Study and Analysis}
\label{sec:ablation}
We perform targeted ablations to disentangle the contributions of our key components: structure-aware support selection and the hybrid control mechanism.

\textbf{Efficacy of Structure-Aware Selection.}
Table~\ref{tab:key_adversary_activation} validates the hypothesis that the selection of controlled agents is as critical as the control mechanism itself.
Randomly selecting a support set (Random-K) yields substantially lower failure rates (28.57\%) compared to our method, proving that sparsity alone is insufficient. While targeting bifurcation points based on topology alone (Topo-only) improves stress, it degrades validity as evidenced by higher AOR and RIR, implying that graph-theoretic analysis may select physically infeasible interventions. Our full pipeline (Topo+Feas.), which filters candidates through semantic feasibility, achieves a favorable balance. It effectively doubles the failure rate of Random-K while maintaining the validity of the generated scenarios.

\textbf{Drift Composition and Topological Anchoring.}
Table~\ref{tab:guidance_pi_ablation} examines the control dynamics. Using only feasibility guidance ($V_{\mathrm{feas}}$) is conservative, while using only adversarial guidance ($V_{\mathrm{adv}}$) is overly aggressive and degrades realism. Combining both creates a necessary regularized objective. Notably, enabling Topological Anchoring provides a performance boost, increasing CFR from 51.43\% to 57.14\%, while preserving realism (REAL increases to 0.7848).
This result confirms that anchoring effectively aligns the reverse-time generation with the structure of the target interaction, preventing the optimizer from collapsing into trivial or unrealistic failure modes.
Fig.~\ref{fig:partial_exp2} visually confirms this behavior, showing that anchoring sharpens the interaction skeleton and concentrates risk on the critical chain of agents.

\subsection{Generalization Across Ego Policies}
\label{sec:generalization}

To assess generalization, we evaluate $E^2$ against three distinct ego policies, specifically rule-based (LG), physics-based (IDM), and hybrid learned (BITS), treating each as a black box. Results in Table~\ref{tab:algo_comparison} suggest that $E^2$ adapts to specific policy failure modes rather than overfitting to a single interaction pattern.

\textbf{Divergent Failure Modes.}
While LG and IDM exhibit identical failure rates (CFR 51.43\%), their collision signatures differ significantly. LG is predominantly vulnerable to side-impact collisions (42.86\%), implying that the adversary exploits the rigid lane-adherence logic of rule-based planners via aggressive cut-ins. Conversely, IDM failures are uniformly distributed across front, side, and rear impacts. Since IDM relies on longitudinal headway, this broad distribution suggests that the adversary overwhelms the simple reactive logic through diverse maneuvers, such as sudden braking or merging.

\begin{table}[t!]
  \caption{\textbf{Closed-loop outcomes across ego policies at fixed intensity ($\eta=2.0$).} Values are averaged over 7 random seeds. Front-col, Side-col, and Rear-col denote the fractions of rollouts with front, side, and rear-impact collisions. All policies use the same observation interface and replanning frequency and are evaluated as black-box controllers.}
  \label{tab:algo_comparison}
  \centering
  \small
  \setlength{\tabcolsep}{3pt}
  \renewcommand{\arraystretch}{1.2}
  \begin{tabular}{lcccccc}
    \toprule
    \makecell{Ego \\ Policy} &
    \makecell{TTC-C} &
    \makecell{CFR \\ (\%)} &
    \makecell{Front-col\\(\%)} &
    \makecell{Side-col\\(\%)} &
    \makecell{Rear-col\\(\%)} &
    \makecell{RelVel \\ ($\mathrm{m/s}$)} \\
    \midrule
    LG & 0.1532 & 51.43 & 11.43 & 42.86 & 28.57 & 2.11 \\
    IDM   & 0.2127 & 51.43 & 25.71 & 24.71 & 25.71 & 2.23 \\
    BITS  & 0.1564 & 25.71 & 17.14 & 20.00 & 17.14 & 19.70 \\
    \bottomrule
  \end{tabular}
\end{table}

\begin{figure}[ht!]
  \begin{center}

    \begin{subfigure}{0.48\columnwidth}
      \centering
      \includegraphics[width=\linewidth]{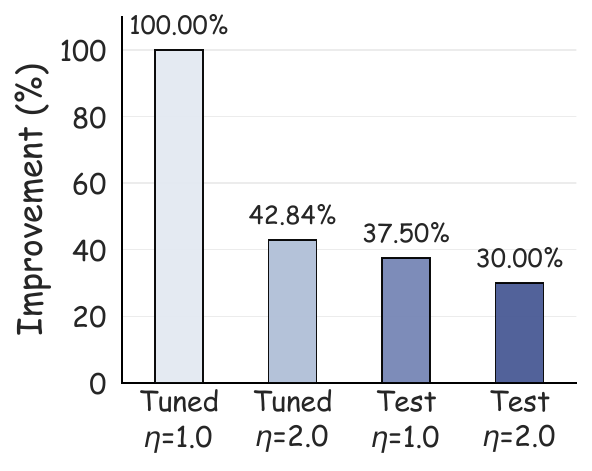}
      \caption*{(a) Failure Rate}
    \end{subfigure}
    \hfill
    \begin{subfigure}{0.48\columnwidth}
      \centering
      \includegraphics[width=\linewidth]{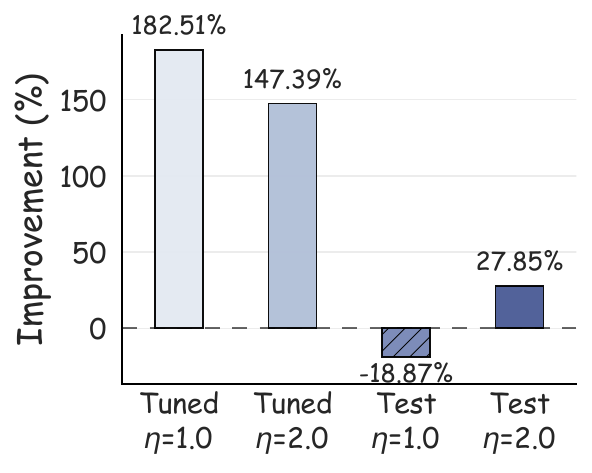}
      \caption*{(b) Min-DTC}
    \end{subfigure}

    \vspace{0.05in}

    \begin{subfigure}{0.48\columnwidth}
      \centering
      \includegraphics[width=\linewidth]{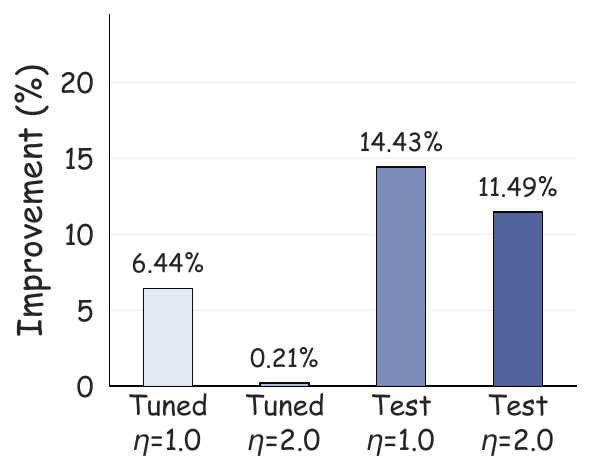}
      \caption*{(c) TTC-cost}
    \end{subfigure}
    \hfill
    \begin{subfigure}{0.48\columnwidth}
      \centering
      \includegraphics[width=\linewidth]{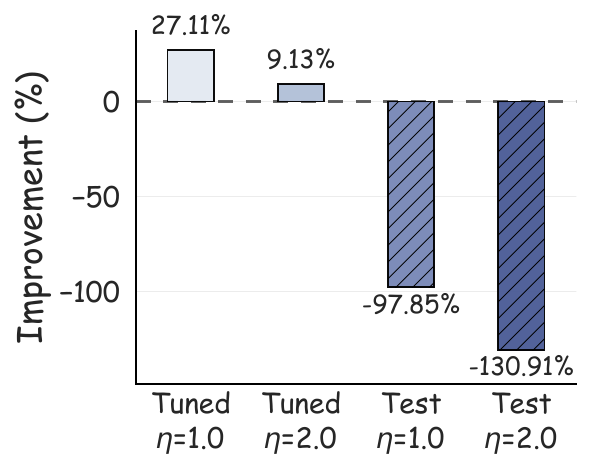}
      \caption*{(d) Mean Acceleration}
    \end{subfigure}
    \caption{\textbf{Policy improvement from adversarial closed-loop fine-tuning.} Bars show the percentage gain over the original lane-graph policy on a tuning subset and a disjoint test subset of the nuScenes validation split, evaluated under matched and increased adversarial intensity ($\eta=1.0,2.0$).}
    \label{fig:policy_learning}
  \end{center}
\end{figure}

\textbf{Robustness-Severity Trade-off.}
The hybrid BITS policy exhibits greater robustness (CFR 25.71\%) but introduces a critical trade-off regarding severity.
In the event of a failure, BITS incurs substantially higher-energy impacts (RelVel $19.70~\mathrm{m/s}$), exceeding the low-speed collisions observed in LG and IDM by an order of magnitude. This observation implies that while BITS effectively manages low-speed interactions, it remains brittle in high-speed scenarios when facing aggressive interventions that the learned policy fails to anticipate.

Overall, these results indicate that $E^2$ serves as a policy-conditioned diagnostic tool. It automatically localizes agent-specific decision boundaries, such as lateral rigidity in planners and high-speed brittleness in learned policies, without requiring manual specification of the failure mode.

\subsection{Impact on Policy Learning}
\label{sec:policy_learning}

Finally, we implement the closed-loop curriculum by employing generated adversarial rollouts as corrective supervision for the lane-graph policy, transforming evaluation into an active learning process. Hyperparameters are fixed via grid search on the tuning split. Fig.~\ref{fig:policy_learning} illustrates the performance gains.

On the tuning split, the policy achieves substantial hazard mitigation, with failure rates dropping by 100.00\% at \(\eta{=}1.0\) and 42.84\% at \(\eta{=}2.0\). This improvement suggests that the adversarial scenarios effectively isolate distinct failure modes, providing informative signals for rapid adaptation.
Crucially, robustness extends to the disjoint test split, where failure rates decrease by 30.00\% to 37.50\% alongside significant TTC improvements. This transferability indicates that the policy acquires a generalized risk representation rather than memorizing specific patterns, allowing it to navigate unseen interactions with greater competence.

However, these gains reveal a trade-off between safety and comfort. Mean acceleration increases on the tuning split (27.11\% at \(\eta{=}1.0\)) but decreases on the test split (-97.85\% and -130.91\%). This implies the agent adopts a dual strategy, executing aggressive maneuvers to evade known threats while becoming more conservative in novel environments to maintain safety margins.

\section{Conclusion}

In this work, we introduce \textbf{Evaluation as Evolution}, a closed-loop framework that unifies adversarial scenario generation, safety evaluation, and policy refinement. Our approach aims to bridge the gap between static validation and active learning by transforming discovered failures into an adaptive source of corrective supervision. The proposed method offers a computationally efficient solution for synthesizing realistic, high-stakes interactions through transport-regularized sparse control over a learned reverse-time SDE prior. 
A notable advantage of $E^{2}$ is its ability to isolate intervention-critical agents via topological bifurcation analysis while maintaining distributional fidelity. Furthermore, Topological Anchoring stabilizes the generation process by injecting interaction-preserving skeletons to guide the model toward ego-critical outcomes without sacrificing realism.

Experimental results on the nuScenes and nuPlan datasets demonstrate the framework's strong zero-shot cross-dataset generalizability and highlight the potential of recycling boundary cases for iterative fine-tuning, leading to substantial gains in policy robustness.
These findings demonstrate that $E^{2}$ serves as a policy-conditioned diagnostic tool, automatically localizing decision boundaries and recovering crucial safety information that is often lost in conventional decoupled pipelines. Future work could explore methods for automatically determining the optimal adversarial intensity and causal templates based on the specific failure modes or the structural characteristics of the interactive environment.

\bibliographystyle{IEEEtran}
\bibliography{references}

\newpage

\end{document}